\documentclass[lettersize,journal]{IEEEtran}
\usepackage{amsmath,amsfonts}
\usepackage{algorithmic}
\usepackage{algorithm}
\usepackage{array}
\usepackage[caption=false,font=normalsize,labelfont=sf,textfont=sf]{subfig}
\usepackage{textcomp}
\usepackage{stfloats}
\usepackage{multirow}
\usepackage{url}
\usepackage{verbatim}
\usepackage{graphicx}
\usepackage{cite}
\usepackage{booktabs}
\usepackage{colortbl}
\usepackage{xcolor} % Add this line to include the xcolor package
\begin{document}

% \title{Fused S$^2$LiNet for HSI and LiDAR Efficient Linear Feature Learning }
\title{HSLiNets: Hyperspectral Image and LiDAR Data Fusion Using Efficient Dual Non-Linear Feature Learning Networks}

\author{{Judy X~Yang,~\IEEEmembership{Graduate Student Member,~IEEE},
        Jing~Wang,
        Chen Hong Sui,
       Zekun~Long,~\IEEEmembership{Graduate Student Member,~IEEE}, and Jun~Zhou,~\IEEEmembership{Senior Member,~IEEE} }
   } % <-this % stops a space
% \thanks{This paper was produced by the IEEE Publication Technology Group. They are in Piscataway, NJ.}% <-this % stops a space
% \thanks{Manuscript received April 19, 2021; revised August 16, 2021.}

 % The paper headers
% \markboth{Journal of \LaTeX\ Class Files,~Vol.~14, No.~8, August~2021}%
% {Shell \MakeLowercase{\textit{et al.}}: A Sample Article Using IEEEtran.cls for IEEE Journals}

% \IEEEpubid{0000--0000/00\$00.00~\copyright~2021 IEEE}
% % Remember, if you use this you must call \IEEEpubidadjcol in the second
% % column for its text to clear the IEEEpubid mark.

\maketitle

\begin{abstract}
The integration of hyperspectral imaging (HSI) and LiDAR data within new non-linear feature spaces offers a promising solution to the challenges posed by the high-dimensionality and redundancy inherent in HSIs. This study introduces a dual non-linear fused space framework that capitalizes on bidirectional reversed convolutional neural network (CNN) pathways, coupled with a specialized spatial analysis block. This approach combines the computational efficiency of CNNs with the adaptability of attention mechanisms, facilitating the effective fusion of spectral and spatial information. The proposed method not only enhances data processing and classification accuracy, but also mitigates the computational burden typically associated with advanced models such as Transformers. Evaluations of the Houston 2013 dataset demonstrate that our approach surpasses existing state-of-the-art models. This advancement underscores the potential of the framework in resource-constrained environments and its significant contributions to the field of remote sensing.
\end{abstract}


\section{Introduction}
 \IEEEPARstart{T}{he} fusion of hyperspectral imaging (HSI) and Light Detection and Ranging (LiDAR) data has garnered considerable attention in remote sensing due to the complementary nature of the information provided by these two technologies~\cite{xu2024joint}. Traditional data fusion methods have laid the foundations for the integration of these complex datasets, but the advent of deep learning techniques has revolutionized the field, allowing for more sophisticated processing and interpretation~\cite{gao2020survey}. In particular, convolutional neural network (CNN) architectures have become instrumental in enhancing the fusion process by effectively extracting and integrating spatial and spectral features~\cite{jing2017adaptive}.

Significant progress has been made with frameworks such as that introduced by Huang et al., which employs dual CNNs to concurrently process spectral-spatial features from HSI and elevation information from LiDAR, setting new benchmarks in the domain~\cite{yang2024lidar}.  Further advances include the development of the Encoder-Decoder network (EndNet)~\cite{hong2020deep} and collaborative contrastive learning (CCL)~\cite{jia2023collaborative},  which have improved the fusion and interpretation of multimodal data, even in scenarios without labeled samples.

Recent research has increasingly focused on integrating attention mechanisms to refine feature extraction and enhance the interaction between HSI and LiDAR data. For example, models such as FusAtNet~\cite{xu2017multisource} and the hybrid 3D/2D CNNs proposed by Falahatnejad et al.~\cite{falahatnejad2022deep} demonstrate the efficacy of employing both self-attention and cross-attention mechanisms to achieve superior classification accuracy.

Although Vision Transformers (ViT)\cite{zhang2024vision} have achieved remarkable success in visual representation learning, particularly in large-scale self-supervised pretraining and downstream tasks, they come with significant computational and memory demands. The SpectralFormer\cite{hong2021spectralformer}, a transformer-based framework for hyperspectral image classification, processes data in a pixel-wise or patch-wise manner and enhances spectral information extraction through cross-layer skip connections. Similarly, the HSI-BERT model~\cite{he2019hsi} uses a bidirectional transformer encoder to effectively capture global dependencies. However, the resource-intensive nature of these models limits their applicability in resource-constrained environments.

To address these challenges, recent research has focused on state-space models (SSMs), which offer improved computational efficiency and the ability to capture long-range dependencies~\cite{gu2021efficiently}. The Mamba model, which uses data-dependent SSM layers, has outperformed Transformer models on various benchmarks while maintaining linear scalability in sequence length~\cite{gu2021efficiently}. This study explores the potential for adapting the Mamba framework for vision applications, proposing a vision backbone architecture that is exclusively based on SSMs without the need for attention mechanisms.

Building on these advancements, our proposed framework, the Fused non-Linear Space, diverges from traditional data fusion approaches by eliminating the need for computationally intensive self-attention modules. Instead, our HSLiNet model leverages bidirectional reversed networks within a novel HSI and LiDAR non-linear fusion framework, achieving comparable modeling capabilities while significantly reducing computational complexity and memory demands. This efficiency makes our framework more suitable for real-time applications and resource-constrained environments, offering a practical alternative to more resource-intensive models such as Mamba and Transformer architectures.

Our research introduces a novel framework, the Fused non-linear Space, which integrates HSI and LiDAR data into an optimized non-linear space, enhancing the extraction and fusion of data features. The primary contributions of this work are as follows.

The main contributions of this article are summarized as follows:
\begin{itemize}
 \item {Feature Selection in Hyperspectral Imaging:} We introduce an innovative Fused non-linear Space framework that not only integrates HSI and LiDAR data but also enhances the selection of the most informative features from hyperspectral images. This approach significantly improves the accuracy and robustness of data representation by focusing on the most relevant spectral bands.
 \item {Reduction in Computational Demands:} Our proposed HSI and LiDAR non-linear Fusion Framework leverages reversed bidirectional networks, which effectively reduce the high computational and memory demands typically associated with deep learning models. This makes the framework highly suitable for real-time applications and resource-constrained environments.
 \item {Performance Comparison with State-of-the-Art Models: }The HSLiNet model, developed within our framework, achieves modeling capabilities that are comparable to advanced models like SpectralFormer, yet without relying on computationally intensive self-attention mechanisms. This results in a model that maintains high performance while offering more efficient resource usage, positioning it as a competitive alternative to existing state-of-the-art methods.
 \end{itemize} 
 
To demonstrate the effectiveness of our HSLiNet model in classifying hyperspectral and LiDAR data, we designed a forward-backward 1D CNN structure as the HSI spectral feature extractor, assisted by LiDAR. Specifically, a sub non-linear spatial network was designed for the fusion of HSI and LiDAR spatial features. This approach resolves the challenges associated with designing separate subnets for HSI and LiDAR data. Experimental results in the Houston data set demonstrate that our proposed fused non-linear model outperforms state-of-the-art (SOTA) models.

\section{Methodology}
In this section, we present the methodology underlying our proposed HSI-LiDAR Fusion non-linear model. We begin by introducing the core algorithms that form the foundation of our approach, followed by a comprehensive description of the model architecture, emphasizing the novel early fusion strategy for integrating HSI and LiDAR data. Finally, we analyze the efficiency of the model, underscoring its computational advantages, and benchmarking its performance against existing methods.

\subsection{HSI-LiDAR non-linear Fusion Model Preliminaries}
We introduce FusedBiRNet, our pioneering model for the analysis and classification of hyperspectral and LiDAR images. Central to our model are the transformation parameters \( A \) and \( B \), which exploit the unique spectral characteristics of hyperspectral data and the spatial properties of HSI and LiDAR. These parameters facilitate a reversed bidirectional processing approach, enriching the model's ability to integrate comprehensive spectral-spatial information effectively.

\subsubsection{Forward and Backward Spectral Dependencies}
In the proposed fused HSLiNet, the parameter \( A \) captures forward spectral dependencies, ensuring a thorough representation of spectral progression. Conversely, the parameter \( B \) seizes backward spectral information, allowing retrospective integration of the spectral data. This bidirectional methodology guarantees a holistic feature representation by amalgamating spectral and spatial information throughout the spectral range. These features are synthesized and classified via mechanisms specifically designed for the discrete nature of hyperspectral and LiDAR data, surpassing the limitations of traditional methods.

\subsection{Proposed Method Overview and Architecture Description}
\begin{figure}[pht]
\centering
\includegraphics[scale=0.42]{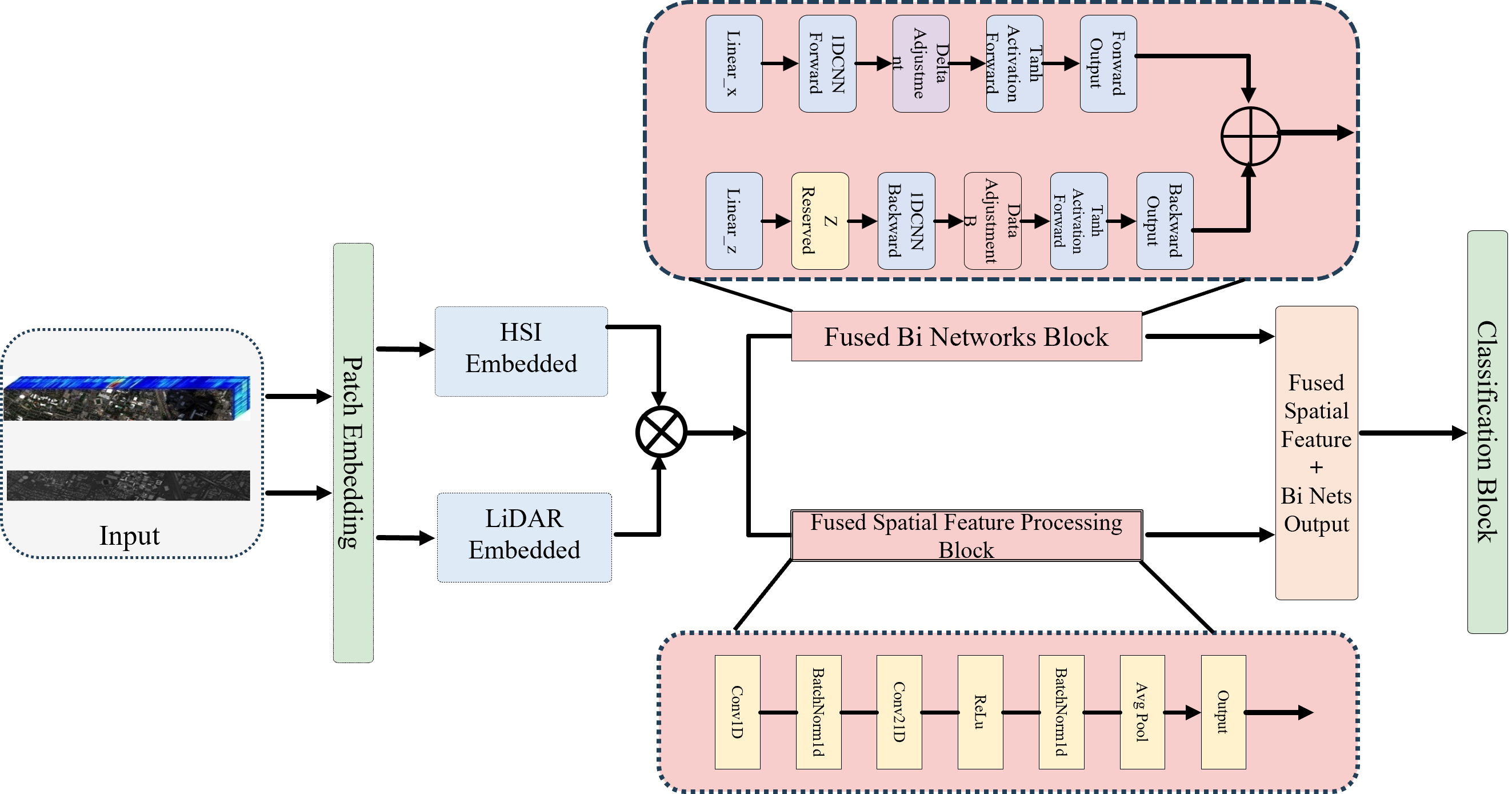}
\caption [width=16cm, height=14cm]{The architectural overview of the Proposed HSI and Lidar Fusion model. The framework consists of four main components: (A) A hyperspectral image patch with dimensions \( p \times p \times \text{CH} \), LiDAR patch with dimension \( p \times p \times \text{1} \); (B) The BiNetworks Block for fused spectral feature and spatial feature processing; (C) Feature fusion Block; (D) The classifier. The process begins by extracting patches that serve as input to the HSI-LiDAR non-linear block. The HSI block includes LiDAR spatial processing stage and HSI unique forward and backward operations, with the concatenation of Binetworks output and LiDAR spatial feature process forming the input to the classifier block.}
\label{hsilidvim}
\end{figure}

Figure~\ref{hsilidvim} illustrates the architecture of the proposed method. The hyperspectral and LiDAR data are initially reshaped and normalized. Post normalization, the data are fused immediately after patch embedding, streamlining the integration process. This fusion enhances the interaction between the spatial features from LiDAR and the spectral features from the hyperspectral data early in the processing pipeline.

\subsection{HSI-LiDAR non-linear Fusion Block}
The HSI-LiDAR Fusion Block, shown in Fig.~\ref{hsilidvim}, is a novel neural network module designed to handle hyperspectral imagery coupled with LiDAR data, characterized by extensive spectral bands and complex spatial structure. This block captures the unique spectral-spatial interplay inherent to the fused hyperspectral and LiDAR data. The Fused Binetworks Block (B-Net) and the LiDAR Spatial Feature Processing Block (S-Block) operate in tandem to process the integrated patches, employing a bidirectional approach where the fused data is processed in both forward and backward directions. Each direction employs a 2D convolution followed by a delta-modulated non-linearity to update the state, encapsulating the spectral dimension's sequential nature.

\begin{equation}
x_{\text{forward}} = f_{\text{activation}}(\text{Conv1d}(x_{\text{proj}}))
\end{equation}

\begin{equation}
x_{\text{backward}} = f_{\text{activation}}(\text{Conv1d}(z_{\text{proj, reversed}}))
\end{equation}

State updates for each direction are modulated by a delta parameter \( \Delta \), which adapts the transformation matrices \( A \) and \( B \) for discrete-time processing:
\begin{equation}
h_{\text{forward}} = \tanh(x_{\text{forward}} + A \cdot \Delta_{\text{expanded}})
\end{equation}

\begin{equation}
h_{\text{backward}} = \tanh(x_{\text{backward}} + B \cdot \Delta_{\text{expanded}})
\end{equation}

The output states from both directions are reduced (e.g., by averaging) and combined to form the final hidden representation \( h \):
\begin{equation}
h_{\text{combined}} = \text{reduce}(h_{\text{forward}}) + \text{reduce}(h_{\text{backward}})
\end{equation}

% HSI+LiDAR Spatial Feature Processing
The fused HSI and LiDAR data undergo spatial processing in the S-Block, where the data patches are first normalized and then transformed through a series of convolutional layers designed to enhance spatial features. The spatial processing involves:
\begin{equation}
hl_{\text{processed}} = \text{ReLU}(\text{BatchNorm}(\text{Conv2d}(hl_{\text{normalized}})))
\end{equation}

% Fusion of Bi-Net and S-Block
The enriched feature sets from both Bi-Net and S-Block are concatenated to form a comprehensive spectral-spatial feature vector:
\begin{equation}
y_{\text{fusion}} = \text{concat}(h_{\text{combined}}, hl_{\text{processed}})
\end{equation}

% Classification
This concatenated feature vector \( y_{\text{fusion}} \) is then passed through additional convolutional layers to further refine the features before being projected to the output dimension and directed towards the classifier block for the classification task:
\begin{equation}
Y = \text{Linear}(\text{Conv1d}(y_{\text{fusion}}))
\end{equation}

where,
\begin{itemize}
    \item \( x(t) \): Fused HSI and LiDAR input data at spectral band \( t \)
    \item \( x_{\text{proj}}, z_{\text{proj}} \): Projected inputs for forward and backward processing
    \item \( x_{\text{forward}}, x_{\text{backward}} \): Hidden states after convolution and activation
    \item \( \Delta_{\text{expanded}} \): Delta parameter expanded for broadcasting
    \item \( h_{\text{forward}}, h_{\text{backward}} \): Updated hidden states after applying \( A \) and \( B \)
    \item \( h_{\text{combined}} \): Combined hidden state from forward and backward paths
    \item \( \text{reduce}() \): Reduction operation 
    \item \( Y \): Final output after linear transformation
    \item \( hl_{\text{normalized}}, hl_{\text{processed}} \): Fused data after normalization and convolutional processing
\end{itemize}

The design of this model differs from traditional models used in text sequence modeling and RGB image token sequence modeling, enhancing the classification accuracy through its specialized approach.

\section{Experiments}
\subsection{Dataset}
\subsubsection{Houston 2013} The 2013 IEEE GRSS Data Fusion dataset~\cite{} features hyperspectral and LiDAR data, including a 144-band hyperspectral image (HSI) spanning 380-1050nm and a LiDAR-derived digital surface model (DSM), both at 2.5 m resolution. The HSI is calibrated for sensor radiation and the DSM measures elevation above sea level. With 15 land cover types, this dataset is suitable for testing band selection and classification techniques, despite urban complexity and HSI noise challenges.
The number of training and test data is extracted based on standard training and test data quantity requirements. 

\subsection{Implementation Details }
Our proposed Fused non-linear Crossed Model, HSLiNet, is implemented using the PyTorch framework. We start by fusing the raw Hyperspectral Imaging (HSI) and LiDAR data before patch embedding. These fused data patches are then processed through the model's forward and backward non-linear direction blocks and spatial blocks. The model is trained using an NVIDIA RTX 3090 GPU with 24 GB of memory, with a batch size set to 32 and an initial learning rate of 0.0001. 

\subsection{Experiment Results}
\begin{figure}[pht]
\centering
\includegraphics[scale=0.35]{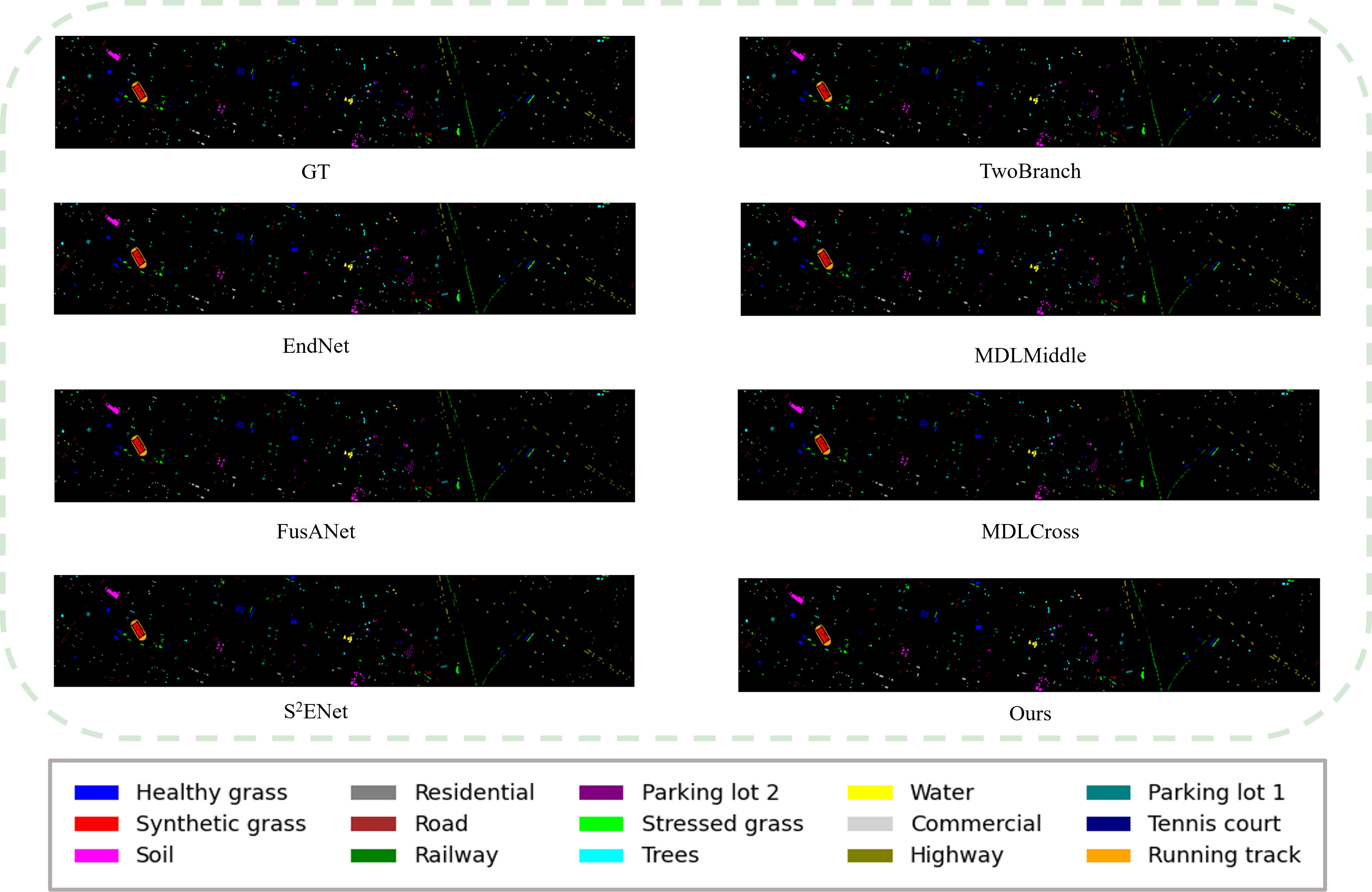} % Update the path to your image
\caption{Visualization and classification maps for the Houston2013 dataset. Ground-truth map, six comparative methods and HSLiNet.}
\label{plots}
\end{figure}

HSLiNet is compared with several representative deep learning-based approaches. The selected methods include:
TwoBranch CNN~\cite{}: A typical two-branched model.
FusAtNet~\cite{xu2017multisource}: A network with a cross-attention mechanism that utilizes LiDAR features to weight HSI features, including a spatial-spectral enhancement module.
EndNet~\cite{hong2020deep}: An encoder-decoder network that uses the reconstruction strategy for feature fusion.
MDL-Middle~\cite{hong2020more}: A baseline CNN model using intermediate fusion.
MDL-Cross~\cite{hong2020more}: A network with a cross-attention mechanism that utilizes LiDAR features to weight HSI features, including a spatial-spectral enhancement module.
$S^2$ENet~\cite{fang2021s2enet}: A model incorporating spatial-spectral enhancement for improved feature extraction and fusion.
These methods are evaluated both visually and quantitatively. Figure \ref{plots} presents visual comparison results, while Table \ref{com} provides quantitative metrics including individual class accuracy, overall accuracy (OA), average accuracy (AA), and Kappa coefficient.

\begin{table*}[!t]
\centering
\caption{Quantitative Comparison of Different Methods on the Houston2013 Dataset. Best Results Are Marked in \textbf{\underline{Bold and Underlined}} and Second Best Results Are Marked in \textbf{Bold}.}\label{tab2}
\begin{tabular}{|l|l|l|l|l|l|l|l|l|}
\hline
No. & Class  & TwoBranch~\cite{xu2017multisource} & EndNet~\cite{hong2020deep} & MDLMiddle~\cite{hong2020more} & FusAtNet~\cite{fang2021s2enet} & MDLCross~\cite{hong2020more} & $S^2$ENet~\cite{fang2021s2enet} &OursHSLiNet \\
\hline
C1 & Healthy grass  & {83.10} & 81.58 & {83.10} & {83.10} &\textbf{83.10}&  82.91&\textbf{\underline{100.00}} \\
C2 & Stressed grass  & 84.10 & 83.65 & 85.06 & 96.05 &84.68&  \textbf{\underline{100.00}}&\textbf{99.34} \\
C3 & Synthetic grass & \textbf{\underline{100.00}} & \textbf{\underline{100.00}} & \textbf{99.60} & \textbf{\underline{100.00}} &\textbf{99.60}&  \textbf{\underline{100.00}}&\textbf{\underline{100.00}} \\
C4 & Trees  & 93.09 & 93.09 & 91.57 & 93.09&92.80& \textbf{96.88}&\textbf{\underline{98.11}} \\
C5 & Soil  & \textbf{\underline{100.00}} & \textbf{99.91} & 98.86 & 99.43&\textbf{\underline{100.00}} & 99.91&\textbf{\underline{100.00}} \\
C6 & Water (182/143) & \textbf{99.30} & 95.10 & \textbf{\underline{100.00}} & \textbf{\underline{100.00}} &\textbf{\underline{100.00}}& \textbf{\underline{100.00}}&\textbf{\underline{100.00}} \\
C7 & Residential & 92.82 & 82.65 & 96.64 & 93.53 &\textbf{\underline{98.51}}& 95.15&\textbf{97.11} \\
C8 & Commercial  & 82.34 & 81.29 & 88.13 & 92.12 &88.89& \textbf{93.92}&\textbf{\underline{97.53}} \\
C9 & Road  & 84.70 & 88.29 & 85.93 & 83.63 &82.06& \textbf{91.31}&\textbf{\underline{97.54}} \\
C10 & Highway  & 65.44 & 89.00 & 74.42 & 64.09 &91.41& \textbf{92.95}&\textbf{\underline{94.21}} \\
C11 & Railway  & 88.24 & 83.78 & 84.54 & 90.13 &\textbf{91.94}&  \textbf{\underline{94.69}}&89.56 \\
C12 & Parking lot 1  & 89.53 & 90.39 & 95.39 & 91.93 &\textbf{\underline{99.14}}& 89.43&\textbf{96.54}\\
C13 & Parking lot 2  & \textbf{92.28} & 82.46 & 87.37 & 88.42 &85.61 & 83.16 &\textbf{\underline{97.89}}\\
C14 & Tennis court  & 96.76 & \textbf{\underline{100.00}} & 95.14 & \textbf{\underline{100.00}}&\textbf{99.95}& \textbf{\underline{100.00}}&\textbf{\underline{100.00}} \\
C15 & Running track  & \textbf{99.79} & 98.10 & \textbf{\underline{100.00}} & 99.15&\textbf{\underline{100.00}}&\textbf{\underline{100.00}}&\textbf{\underline{100.00}} \\
\hline
 & OA & 87.98 & 88.52 & 89.55 & 89.98 &91.99&  \textbf{94.19}&\textbf{\underline{96.68}} \\
 & AA & 90.11 & 89.95 & 91.05 & 91.65 &92.91&  \textbf{94.69}&\textbf{\underline{97.32}} \\
 & Kappa & 86.98 & 87.59 & 88.71 & 89.13 &91.33& \textbf{93.69}&\textbf{\underline{96.39}} \\
\hline
\end{tabular}
\label{com}
\end{table*}

The table presents a quantitative comparison of different methods in the Houston2013 dataset, with the best results for each class marked in \textbf{\underline{Bold and Underlined}} and the second-best results in \textbf{Bold}.

For the Healthy Grass class (C1), HSLiNet achieved the highest accuracy with a perfect score of 100.00, outperforming other methods that scored around 83.10. For Stressed Grass (C2), HSLiNet again achieved a near-perfect score of 99.34, closely following MDLCross, which scored 100.00. Synthetic grass (C3) showed perfect results for most methods, each scoring 100.00, except MDLMiddle and MDLCross, which scored 99.60.

In the Trees category (C4), HSLiNet achieved the highest accuracy with a score of 98.11, while other methods scored between 91.57 and 96.88. For the Soil category (C5), several methods, including HSLiNet, achieved perfect scores. In the category Water (C6), multiple methods, including HSLiNet, achieved perfect scores, while TwoBranch and EndNet scored slightly lower.

HSLiNet led the Residential (C7) category with a score of 97.11, surpassing other methods that scored between 82.65 and 98.51. For the Commercial (C8) category, HSLiNet performed the best with a score of 97.53, followed by FusAtNet at 93.92, with other methods scoring between 81.29 and 92.12. In the category Road (C9), HSLiNet again led with a score of 97.54, outperforms

\subsection{Ablation Study}
The ablation study dissects the novel bidirectional processing mechanism of the SS non-linear model, underscoring the importance of forward and backward pathways, along with the spatial processing block, in achieving precision in classification. Through iterative removal of these key components, we evaluate their individual impact on the model's performance. The ablation is rigorously performed on the Houston 2013 dataset, providing a detailed investigation into the constituent elements that enhance the classification efficacy of the SS non-linear Model.

Table~\ref{hyper_ab} encapsulates the variations of the SS non-linear Model architecture and their corresponding classification outcomes, affirming the indispensable role of each component. The findings delineate how the concerted operation of bidirectional processing and spatial analysis propels the model to superior classification results. Specifically, the fully integrated model (\underline{\textbf{Method1}}) exhibits the highest performance in overall accuracy (OA), average accuracy (AA) and Kappa metrics, underscoring the critical synergy between forward, reversed, and spatial components to achieve optimal classification accuracy.
In the category Highway (C10), HSLiNet led with a score of 94.21, while other methods scored between 64.09 and 92.95. For the Railway (C11) category, S$^2$ENet achieved the highest score of 94.69, followed by MDLCross, with HSLiNet scoring 89.56. In the category Parking Lot 1 (C12), MDLCross performed the best with a score of 99.14, followed by HSLiNet at 96.54.

In the category Parking Lot 2 (C13), HSLiNet led with a score of 97.89, with other methods scoring between 82.46 and 92.28. For the Tennis Court (C14) category, HSLiNet and several other methods achieved perfect scores. In the Running Track (C15) category, HSLiNet along with other methods achieved perfect scores.

In general, HSLiNet demonstrated the highest overall accuracy (OA) at 96. 68\%, the highest average accuracy (AA) at 97. 32\%, and the highest Kappa coefficient at 96.39\%, significantly outperforming other methods on various metrics in the Houston2013 dataset.

\begin{table}[!t]
\centering
\caption{Different Methods for ABLATION ANALYSIS Based on Houston 2013 Data (HSI+LiDAR)}
\label{hyper_ab}
\begin{tabular}{|l|c|c|c|c|c|c|}
\hline
Methods & Forward & Reversed & Spatial & OA & AA & Kappa \\
\hline
\underline{\textbf{Model1}} & $\checkmark$ & $\checkmark$ & $\checkmark$ & \underline{\textbf{0.9668}} & \underline{\textbf{0.9722}} & \underline{\textbf{0.9639}} \\
Model2 & $\checkmark$ & $\checkmark$ & $\times$ & 0.9585 & 0.9663 & 0.9550 \\
Model3 & $\checkmark$ & $\times$ & $\checkmark$ & 0.9661 & 0.9715 & 0.9632 \\
Model4 & $\times$ & $\checkmark$ & $\checkmark$ & 0.9615 & 0.9691 & 0.9582 \\
Model5 & $\times$ & $\times$ & $\checkmark$ & 0.9342 & 0.9434 & 0.9286 \\
\hline
\end{tabular}
\end{table}

Table~\ref{tab:metrics} presents the performance metrics for different architectures using HSI+LiDAR, HSI, and LiDAR data. Model 1, which integrates both HSI and LiDAR, demonstrates superior performance with the highest Overall Accuracy (OA) of 0.9736, the average accuracy (AA) of 0.9786, and Kappa of 0.9713. This highlights the effectiveness of combining HSI and LiDAR data, as it significantly enhances classification accuracy compared to using HSI or LiDAR alone. The results underscore the importance of multimodal data fusion in improving remote sensing tasks.
\begin{table}[!t]
\centering
\caption{Architecture Metrics for HSI+LiDAR, HSI, and LiDAR Based On Training Model (100Epochs)}
\label{tab:metrics}
\begin{tabular}{|l|l|c|c|c|}
\hline
Architecture & Metrics & HSI+LiDAR & HSI & LiDAR \\ 
\hline
\multirow{3}{*}{{Model 1}} & OA & \underline{\textbf{0.9668}} &  0.9613 & 0.2738 \\ 
 & AA & \underline{\textbf{0.9722}} & 0.9685 & 0.2833 \\ 
 & Kappa & \underline{\textbf{0.9639}} &  0.9583 & 0.2208 \\ 
\hline
\multirow{3}{*}{Model 2} & OA & 0.9538 & 0.9430 & 0.2523 \\ 
 & AA & 0.9609 &  0.9535 & 0.2657 \\ 
 & Kappa & 0.9498 &  0.9381 & 0.2208 \\ 
\hline
\end{tabular}
\end{table}

\begin{figure}[pht]
\centering
\includegraphics[width=8cm, height=6cm]{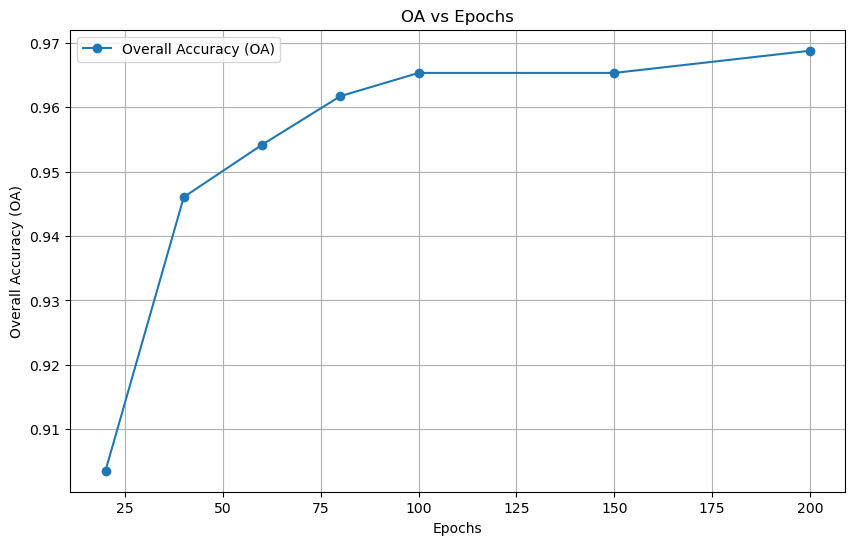} % Update the path to your image
\caption{OA and Epoch relationship based on different Training loops }
\label{ep1}
\end{figure}

We trained the model based on epochs 20, 40, 60, 80, 100, 150, and 200, and we obtained a relationship chart between the epoch and OA in \ref{ep1}. The testing time is always 0.48 seconds, the training time is increasing with the epoch. 

These experiments can be implemented on a computer with less memory than other models.

\section{Conclusion}

In this study, we presented a novel dual fused non-linear framework, the Fused non-linear Space, which integrates hyperspectral imaging (HSI) and light detection and range (LiDAR) data into an optimized non-linear space. This innovative approach improves feature extraction and fusion accuracy while addressing the high computational demands typically associated with deep learning models.

Our framework leverages bidirectional reversed networks to efficiently fuse HSI and LiDAR features, significantly improving classification accuracy. The proposed Hi-Vim model, which operates without a self-attention module, achieves modeling power comparable to that of state-of-the-art methods such as SpectralFormer but with much lower computational complexity and linear memory usage.

The experimental results in the Houston2013 dataset demonstrate the superior performance of our model. The HSLiNets fused consistently outperform other methods, achieving the highest scores in general accuracy (OA), average accuracy (AA) and kapa. These results highlight the effectiveness of our approach in capturing intricate details of multimodal data and set new benchmarks for remote sensing applications.

In summary, our proposed HSLiNets not only address the computational challenges in hyperspectral and LiDAR data fusion but also introduce highly efficient computing for remote sensing tasks.

\bibliographystyle{IEEEtran}

\bibliography{IEEEabrv, Reference.bib}

\end{document}